\documentclass{article}
\usepackage{arxiv}
\usepackage{times}
\usepackage{latexsym}
\usepackage{amsmath}
\usepackage{amssymb}
\usepackage{amsfonts}
\usepackage{booktabs}
\usepackage{threeparttable}
\usepackage{url}
\usepackage{graphicx}
\usepackage{epsfig}
\usepackage{cite}
\usepackage{multirow}
\usepackage{comment}

\usepackage[caption=false, font=footnotesize]{subfig}

\graphicspath{{./figures/}}

\title{Text Classification based on Multi-granularity Attention Hybrid Neural Network}

\author{ Zhenyu~Liu \\
  Department of Science and Technology Teaching\\
  China University of Political Science and Law\\
  Beijing 100088\\
   \And
 Chaohong~Lu \\
  School of Computer and Science Technology\\
  University of Science and Technique of China\\
  Hefei, Anhui 230027\\
  \And
 Haiwei~Huang \\
  School of Computer and Science Technology\\
  University of Science and Technique of China\\
  Hefei, Anhui 230027\\
  \And
Shengfei~Lyu \\
  School of Computer and Science Technology\\
  University of Science and Technique of China\\
  Hefei, Anhui 230027\\
  \And
  Zhenchao Tao \\
  The First Affiliated Hospital \\
  University of Science and Technique of China\\
  Hefei, Anhui 230027\\
}

\begin{document}
\maketitle

\begin{abstract}
Neural network-based approaches have become the driven forces for Natural Language Processing (NLP) tasks. Conventionally, there are two mainstream neural architectures for NLP tasks: the recurrent neural network (RNN) and the convolution neural network (ConvNet). RNNs are good at modeling long-term dependencies over input texts, but preclude parallel computation. ConvNets do not have memory capability and it has to model sequential data as un-ordered features. Therefore, ConvNets fail to learn sequential dependencies over the input texts, but it is able to carry out high-efficient parallel computation. As each neural architecture, such as RNN and ConvNets, has its own pro and con, integration of different architectures is assumed to be able to enrich the semantic representation of texts, thus enhance the performance of NLP tasks. However, few investigation explores the reconciliation of these seemingly incompatible architectures. To address this issue, we propose a hybrid architecture based on a novel hierarchical multi-granularity attention mechanism, named Multi-granularity Attention-based Hybrid Neural Network (MahNN). The attention mechanism is to assign different weights to different parts of the input sequence to increase the computation efficiency and performance of neural models. In MahNN, two types of attentions are introduced: the syntactical attention and the semantical attention. The syntactical attention computes the importance of the syntactic elements (such as words or sentence) at the lower symbolic level and the semantical attention is used to compute the importance of the embedded space dimension corresponding to the upper latent semantics. We adopt the text classification as an exemplifying way to illustrate the ability of MahNN to understand texts. The experimental results on a variety of datasets demonstrate that MahNN outperforms most of the state-of-the-arts for text classification. 
\end{abstract}

\section{Introduction}

Nature language understanding plays an critical role in machine
intelligence and it includes many challenging NLP tasks such as
reading comprehension \cite{luong2015effective}, machine translation
\cite{SutskeverVL14}, question answering \cite{BordesCW14} and etc..
Amongst a wide spectrum of NLP tasks, text
classification\cite{Jiang2018LatentTT} is considered as the
foundation for its measuring the semantic similarities between
texts. Traditional machine learning methods employ hand-crafted
features to model the statistical properties of syntactical elements
(usually words), which are further fed into the classification
algorithms such as k-Nearest Neighbor (k-NN), Random Forests,
Support Vector Machines (SVM), or its probabilistic versions
\cite{liu2017predictive,chen-2009-pcvm,chen-2014-epcvm}. However,
such hand-crafted features often suffered from the loss of semantic
information and scalability. To solve the drawbacks of the
hand-crafted features, automatic learning of representation using
the neural networks was introduced into NLP fields. Word embedding
is a foretype of automatic representation learning
\cite{liu2017revisit,xu2016improve}, which outperforms the
traditional methods for alleviating the sparsity problem and
enhancing the semantic representation.

In recent years, the NLP community has conducted extensive
investigations on the neural-based approaches
\cite{bengio2003neural, le2014distributed}. There exist a diversity
of deep neural network architectures with different modeling
capabilities. The RNN is a widely-used neural network architecture
for NLP tasks owing to its capability to model sequences with
long-term dependencies \cite{tang2015document}. When modeling texts,
a RNN sequentially processes word by word and generates a hidden
state at each time step to represent all previous words. However,
although the purpose of RNNs is to capture the long-term
dependencies, theoretical and empirical studies have revealed that
it is difficult for RNNs to learn very long-term information. To
address this problem, the long short-term memory network (LSTM)
\cite{hochreiter1997long} and other variants such as gated recurrent
unit (GRU) \cite{cho2014learning}, simple recurrent unit (SRU)
\cite{lei2017training} were proposed for better remembering and
memory accesses. Another roadblock for RNNs is that when they are
used to process a long sequence, the latest information is more
dominant than the earlier one, however, which might be the real
significant part of the sequence. In fact, the most important
information can appear anywhere in a sequence rather than at the
end. Consequently, some researchers proposed to assign the same
weight to all hidden states and average the hidden states of all
time steps to equally spread the focus to all the sequence.

Inspired by the biological ability to focus on the most important
information and ignore the irrelevant ones, the attention mechanism
was introduced to assign different weights to the elements at
different positions in a sequence and select the informative ones
for the downstream tasks \cite{bahdanau2014neural}. Nowadays, the
attention mechanism has become an integral part of sequence
modeling, especially with RNNs \cite{luong2015effective}. The
attention mechanism enables RNNs to maintain a variable-length
memory and compute the outputs based on the importance weights of
different parts in a sequence. The attention mechanism has been
empirically proven to be effective in NLP tasks such as neural
machine translation \cite{cho2014learning}. However, the attention
mechanism cannot capture the relationships between words and the
word ordering information, which contains important semantic
information for downstream tasks. Taking the sentences
``\textit{Tina likes Bob.}" and ``\textit{Bob likes Tina.}" as
examples, the weighted sum of their hidden states encoded by RNN are
almost the same. Nevertheless, the two sentences have different
meanings.


The ConvNet is another widely-adopted neural architecture for NLP tasks. The modeling power of ConvNets relies on four key factors: local connections, shared weight, pooling and multi-layers. The fundamental assumption behind the ConvNet approaches is that locally grouped data in natural signals are often high correlated and the compositional hierarchies in natural signals can be exploited by the stacked convolutional layers. As a result, ConvNets have been believed to be good at extracting informative semantic representations from the salient N-gram features of input word sequences by utilizing convolutional filters in a parallel way. For the above example, 2-gram features of `` \textit{Tina likes}" and ``\textit{likes Bob}" that contain the word ordering information can be captured by ConvNets. These features are more representative for the original sentence than the weighted sum of the hidden states. Therefore, ConvNets have been employed for a variety of NLP tasks and achieved impressive results in sentence modeling \cite{kalchbrenner2014convolutional}, semantic parsing \cite{yih2014semantic}, and text classification \cite{kim2014convolutional}. Moreover, ConvNets can operate on different levels of lexical structures such as characters, words, sentences, or even the whole document. For instance, some research has shown that the character-level text classification using ConvNets can achieve competitive results in comparison with the state-of-the-arts \cite{zhang2015character,xu-etal-2016-improve}. However, basic ConvNets apply a fixed-width window to slide over the input sequences, which limits the created representations to local semantic pattern, failing to capture long-term dependencies.

To take full advantage of both the ConvNet and the RNN, and
complement the superiorities of different neural architectures,
researchers explored to introduce the hybrid structure of the
ConvNets and the RNNs. For instance, the recurrent convolutional
neural network \cite{lai2015recurrent} proposed a recurrent
structure of convolutional filters to enhance the contextual
modeling ability to avoid the problem of fixed-width sliding
windows. This work also claimed to apply a max-pooling layer to
automatically determine the key components for text classification.
However, even though this approach managed to reduce noise by
replacing the fixed-width sliding window of ConvNets with a
recurrent mechanism, it still depend on the max-pooling to determine
the discriminative features and lacks the mechanism to selectively
choose the dominant component as the attention mechanism can do.
Similarly, Wang et al. proposed the convolutional recurrent neural
network\cite{wang2017hybrid} that stacked four types of neural
layers: word embedding, Bidirectional RNN layer, convolutional
layer, and max-pooling layer. This approach functions very similarly
to the one in \cite{lai2015recurrent}, but with disparate
applications in sentence classification and answer selection. Also,
this work bypassed the attention mechanism when integrating the
ConvNet and the RNN structures.

As discussed above, any neural architecture has its own pros and
cons, it is reasonable to conjecture that consistently combing
different architectures can benefit extracting of different aspects
of linguistic information from texts. However, to the best of our
knowledge, there are still no efforts in integrating entirely the
ConvNet, RNN and attention architectures. Inspired by proposition by
LeCun et al. \cite{lecun2015deep}, we hypothesize that the attention
mechanism can function as an adhesive that seamlessly integrate the
ConvNet and the RNN architecture, where the RNN layer is used to
represent the input word sequences and the ConvNet layer is used for
classification. Furthermore, we assume that, besides attending to
elements (words as a typical example) at syntactical or symbolic
level, coarser-grained attentions at the hidden state vectorial
space can improve the local N-gram coherence for ConvNets, as the
attentions on hidden state vectors can select the salient dimensions
that represent most informative latent semantics, hence reducing the
noise perturbation to the ConvNet layer and enhancing the
classification performance.

Based on the above motivations, we propose a hybrid architecture
based on a novel hierarchical multi-granularity attention mechanism,
named Multi-granularity Attention-based Hybrid Neural Network
(MahNN). In MahNN, two types of attentions are introduced: the
syntactical attention and the semantical attention. The syntactical
attention computes the importance of the syntactic elements (such as
words or sentence) at the lower symbolic level and the semantical
attention is used to compute the importance of the embedded space
dimension corresponding to the upper latent semantics. We adopt the
text classification as an exemplifying way to illustrate the ability
of MahNN to understand texts. The experimental results on a variety
of datasets demonstrate that MahNN outperforms most of the state-of-the-arts for text classification.

\begin{figure}[t!]
\centering
\includegraphics[width=\linewidth]{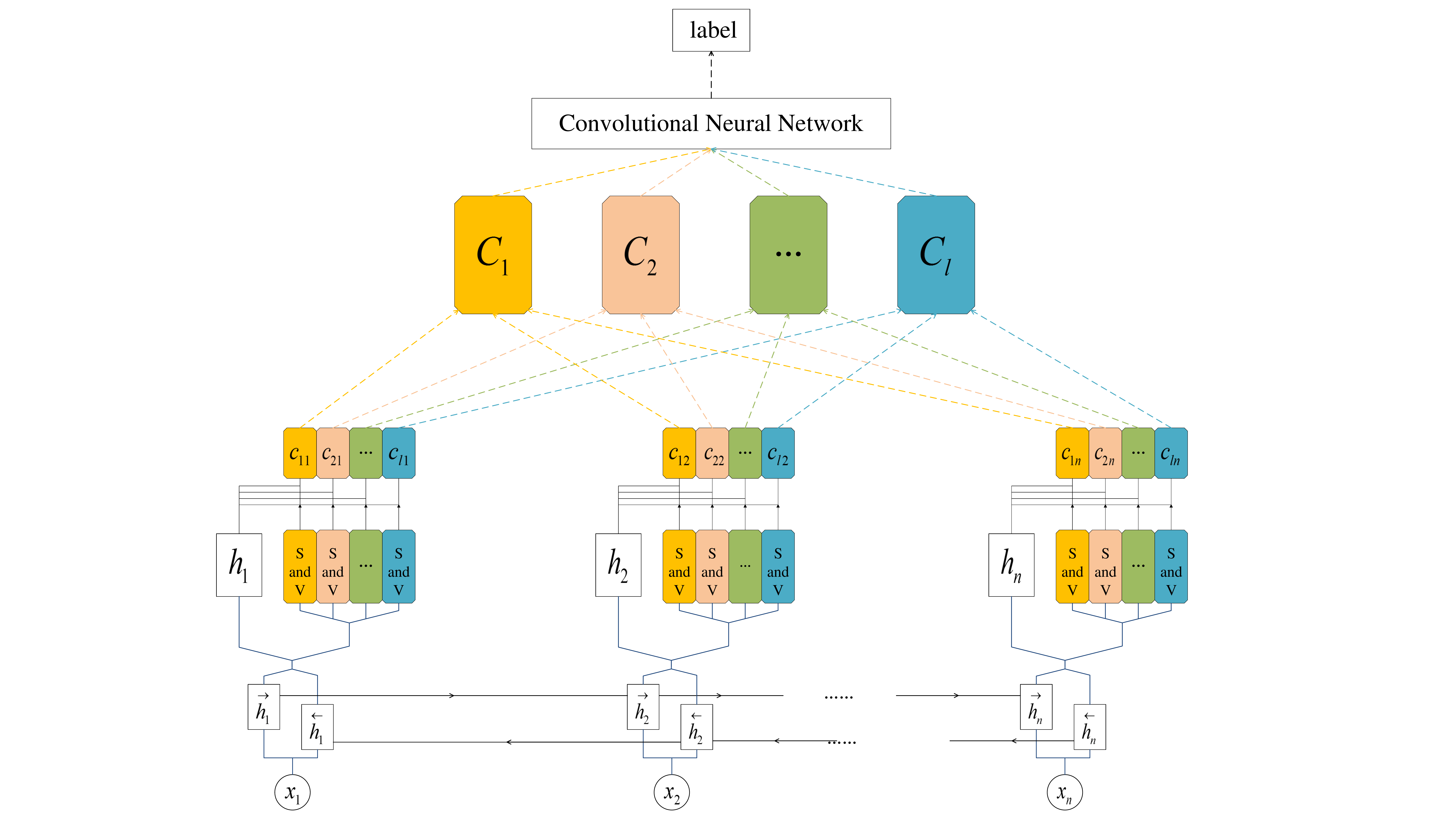}
\caption{The structure of the Attention-based Multichannel Convolutional Neural Network. S and V denote the syntactical attention and the semantical attention, respectively. Blocks of the same color are merged into one channel}
\label{fig:model}
\end{figure}

The main contributions of our work are listed as follows:

\begin{enumerate}
    \item We propose a hybrid neural architecture MahNN that, for the first time, seamlessly integrate the RNN architecture and the ConvNet with an attention mechanism. In this architecture, the different neural structure each learns a different aspect of semantic information from the linguistic structures and collectively strengthen the power of semantical understanding of texts.

    \item we introduce a novel hierarchical multi-granularity attention mechanism, which includes the syntactical attention and the semantical attention. The syntactical attention and the semantical attention compute the importance weights at the lower symbolic level and the upper latent semantics level, respectively. This coarser-grained attention mechanism helps to learn semantic representations more precisely.

\end{enumerate}

This paper is organized as follows. Section \ref{sec:Related work}
introduces the related work about ConvNet and attention mechanisms.
Section \ref{sec:Method} introduces the proposed MahNN in detail.
And Section \ref{sec:Experiment} introduces datasets, baselines,
experiments, and analysis. Finally, Section \ref{sec:Conclusion}
concludes this paper.

\section{Related work}
\label{sec:Related work}

Most of the previous work has exploited deep learning to deal with
NLP tasks, including learning distributed representations of words,
sentences or documents
\cite{mikolov2013distributed,le2014distributed,kalchbrenner2014convolutional,wang2017hybrid}
and text classification
\cite{zhang2015character,yang2016hierarchical,joulin2016bag,lai2015recurrent},
etc.


A ConvNet architecture \cite{kim2014convolutional} was proposed with
multiple filters to capture local correlations followed by
max-pooling operation to extract dominant features. This
architecture performs well on text classification with a few
parameters. The case of using character-level ConvNet was explored
for text classification without word embedding
\cite{zhang2015character} and in this work language was regarded as
a kind of signal. Based on character-level representations, very
deep convolutional networks (VDConvNet)\cite{conneau2016very} were
applied to text classification which is up to 29 convolutional
layers much larger than 1 layer used by \cite{kim2014convolutional}.
To capture word correlations of different sizes, a dynamic
$k$-max-pooling method, a global pooling operation over linear
sequences, was proposed to keep features
better\cite{kalchbrenner2014convolutional}. Tree-structured
sentences were also explored convolutional
models\cite{mou2015discriminative}. Multichannel variable-size
convolution neural network (MVConvNet) \cite{yin2016multichannel}
combined diverse versions of pre-trained word embedding and used
varied-size convolution filters to extract features.

A RNN is often employed to process temporal sequences. In addition
to RNN, there are several approaches for sequences learning, such as
echo state network and learning in the model space
\cite{li2018symbolic,gong2018sequential,chen2014cognitive}. In the
learning in the model space, it transforms the original temporal
series to an echo state network (ESN), and calculates the `distance'
between ESNs  \cite{ChenTRY14,chen2013model}. Therefore, the
distance based learning algorithms could be employed in the ESN
space \cite{gong2016model}. Chen et al. \cite{chen2015model}
investigated the trade-off between the representation and
discrimination abilities. Gong et al. proposed the multi-objective
version for learning in the model space
\cite{gong2018multiobjective}.

The other popular RNN architecture is able to deal with input
sequences of varied length and capture long-term dependencies. Gated
recurrent neural network (GRU) \cite{chung2014empirical} was
proposed to model sequences. As a similar work, GRU was applied to
model documents\cite{tang2015document}. Their works show that GRU
has the ability to encode relations between sentences in a document.
To improve the performance of GRU on large scale text, hierarchical
attention networks (HAN)\cite{yang2016hierarchical} was proposed.
HAN has a hierarchical structure including word encoder and sentence
encoder with two levels of attention mechanisms.

As an auxiliary way to select inputs, attention mechanism is widely
adopted in deep learning recently due to its flexibility in modeling
dependencies and parallelized calculation. The attention mechanism
was introduced to improve encoder-decoder based neural machine
translation \cite{bahdanau2014neural}. It allows a model to
automatically search for parts of elements that are related to the
target word. As an extension, global attention and local attention
\cite{luong2015effective} were proposed to deal with machine
translation and their alignment visualizations proved the ability to
learn dependencies. In HAN \cite{yang2016hierarchical}, hierarchical
attention was used to generate document-level representations from
word-level representations and sentence-level representations. This
architecture simply sets a trainable context vector as a high-level
representation of a fixed query. This way may be unsuitable because
the same words may count differently in varied contexts. In a recent
work \cite{vaswani2017attention}, the calculation of attention
mechanism was generalized into Q-K-V\footnote{Q-K-V denotes query,
key and value respectively.} form.

\section{Multi-granularity Attention-based Hybrid Neural Network}
\label{sec:Method}

The MahNN architecture is demonstrated in Fig.\ref{fig:model}. It consists of three parts: bi-directional long short-term memory (Bi-LSTM), attention layer and convolutional neural network (ConvNet). The following sections describe how we utilize Bi-LSTM to generate the syntactical attention and the semantical attention, and form multichannel for ConvNet.

\subsection{Long Short-Term Memory Network}

In many NLP tasks, RNN processes word embedding for texts of variable length and generates a hidden state $ h_{t} $ in $t$ time step by recursively transforming the previous hidden state $h_{t-1}$ and the current input vector $ x_t $.

\begin{equation}
{h_t} = f(W \cdot [{h_{t - 1}},{x_t}] + b),
\end{equation}

where $ W \in {\mathbb{R}^{{l_h} \times \left( {{l_h} + {l_i}} \right)}} $ , $ b \in {\mathbb{R}^{{l_h}}} $ , $ l_h $ and $ l_i $ are dimensions of hidden state and input vector respectively, and $ f\left(  \cdot  \right) $ represents activation function such as $ tanh\left(  \cdot  \right) $. However, standard RNN is not a preferable choice for researchers due to the problem of gradient exploding or vanishing \cite{bengio1994learning}. To address this problem, the long short-term memory network (LSTM) was introduced and obtained remarkable performance.

As a variant of RNNs, the LSTM architecture has a range of tandem modules whose parameters are shared. At $t$ time step, the hidden state $ h_t $ is controlled by the previous hidden state $h_{t-1} $, input $ x_t $, forget gate $ f_t $, input gate $ i_t $ and output gate $ o_t $. These gates identify the way of updating the current memory cell $ c_t $ and the current hidden state $ h_t$. The LSTM transition function can be summarized by the following equations:
\begin{equation}
\begin{split}
{f_t} &= \sigma ({W_f} \cdot \left[ {{h_{t - 1}},{x_t}} \right] + {b_f}), \\
{i_t} &= \sigma ({W_i} \cdot \left[ {{h_{t - 1}},{x_t}} \right] + {b_i}), \\
{o_t} &= \sigma ({W_o} \cdot \left[ {{h_{t - 1}},{x_t}} \right] + {b_o}), \\
\mathop {{C_t}}\limits^ \sim   &= \tanh ({W_C} \cdot \left[ {{h_{t - 1}},{x_t}} \right] + {b_C}), \\
{C_t} &= {f_t} \odot {C_{t - 1}} + {i_t} \odot \mathop {{C_t}}\limits^ \sim,  \\
{h_t} &= {o_t} \odot \tanh ({C_t}). \\
\end{split}
\end{equation}

Here, $ \sigma $ is the logistic sigmoid function that has the
domain of all real numbers, with return value ranging from 0 to 1. $
tanh $ denotes the hyperbolic tangent function with return value
ranging from -1 to 1. Intuitively, the forget gate $ f_t $ controls
the extent to which the previous cell state $ C_{t-1} $ remains in
the cell. The input gate $ i_t $ controls the extent to which a new
input flows into the cell. The output gate $ o_t $ controls the
extent to which the cell state $ C_t $ is used to compute the
current hidden state $ h_t $. The existence of those gates enables
LSTM to capture long-term dependencies  when dealing with
time-series data.


Though unidirectional LSTM includes an unbounded sentence history in
theory, it is still constrained since the hidden state of each time
step fails to model future words of a sentence. Therefore, Bi-LSTM
provides a way to include both previous and future context by
applying one LSTM to process sentence forward and another LSTM to
process sentence backward.

Given a sentence of $n$ words $\{ w_i \}_{i=1}^n$, we first transfer
the one-hot vector $ w_i $ into a dense vector $ x_i $ through an
embedding matrix $ W_e $ with the equation $ {x_i} = {W_e}{w_i}  $.
We use Bi-LSTM to get the annotations of words by processing
sentence from both directions. Bi-LSTM contains the backward $
\overleftarrow {LSTM} $ that reads the sentence from $ x_n $ to $
x_i $ and a forward $\overrightarrow {LSTM}$ which reads from $ x_1
$ to $ x_i $:
\begin{equation}
\begin{split}
{x_i} &= {W_e}{w_i},i \in \left[ {1,n} \right], \\
\mathop {{h_i}}\limits^ \to &= \overrightarrow {LSTM} ({x_i}),i \in \left[ {1,n} \right], \\
\mathop {{h_i}}\limits^ \leftarrow &= \overleftarrow {LSTM} ({x_i}),i \in \left[ {1,n} \right].
\end{split}
\end{equation}

At $i$ time step, we obtain the forward hidden state $\mathop
{{h_i}}\limits^ \to $ which stores previous information and the
backward hidden state $ \mathop {{h_i}}\limits^ \leftarrow $ which
stores future information. $ {h_i} = [{\mathop {{h_i}}\limits^
\to},{\mathop {{h_i}}\limits^ \leftarrow}] $ is a summary of the
sentence centered around $ w_i $.

\subsection{Hierarchical Multi-granularity Attentions}

For the NLP tasks such as text classification and sentiment
analysis, different words contribute unequally to the representation
of a sentence. The attention mechanism can be used to reflect the
importance weight of the input element so that the relevant element
contributes significantly to the merged output. Although the
attention mechanism is able to model dependencies flexibly, it is
still a crude process because of the loss of latent semantic
information. We apply attention mechanisms to the hidden states of
Bi-LSTM and splice them into a matrix.

Taking the form of the matrix rather than a weighted sum of vectors
will keep the order information. Furthermore, by applying the syntactical
attention and the semantical attention, we could obtain several
matrices and take them as multichannel for inputs of ConvNet.

\subsubsection{Syntactical Attention Mechanism}

We introduce the syntactical attention to calculate the importance
weights of all input elements. $ M $ is the association matrix that
represents the association among words in texts. The element of the
$ i $-th row and the $ j $-th column of $ M $ represents the degree
of association between the $ i $-th word and the $ j $-th word. We
will set $ L $ channel mask matrices $ V $ if we need $ L $
channels. In the $ l $-th channel, $ M_{l_{i,j}} $ is calculated as
follows:
\begin{equation}
M_{l_{i,j}} = \tanh ( [{h_i} , {W_l} \cdot {h_j} ]  + b_l),
\end{equation}
The $ i $-th channel mask matrix is defined as follows:
\begin{equation}
V_{{l_{i,j}}} \sim B(1,{p_l}),i \in [1,n],j \in [1,n],
\end{equation}
That means each element of $ V_l $ obeys binomial distribution.
Given $ M_{l_{i,j}} $ and $ V_{{l_{i,j}}} $, the $ i $-th channel is computed as follows:
\begin{equation}
{A_l} = {M_l} \otimes {V_l},
\end{equation}
    \begin{equation}\label{sum_along}
{s_{lk}} = \sum\nolimits_x {{A_{l_{xk}}}},
\end{equation}
\begin{equation}
p_k = \\
\begin{cases}
-99999, \qquad  if\   x_k\  is\ from\ pad & \\
0,  \qquad \qquad       otherwise\ & \\
\end{cases}
\end{equation}
\begin{equation}
score_{lk} = p_k + s_{lk},
\end{equation}
\begin{equation}\label{syntactical}
a_{lk} = \frac{{\exp (scor{e_{lk}})}}{{\sum\nolimits_{i = 1}^n {\exp (scor{e_{li}})} }},
\end{equation}

\begin{figure}
\centering
\includegraphics[width=0.8\linewidth]{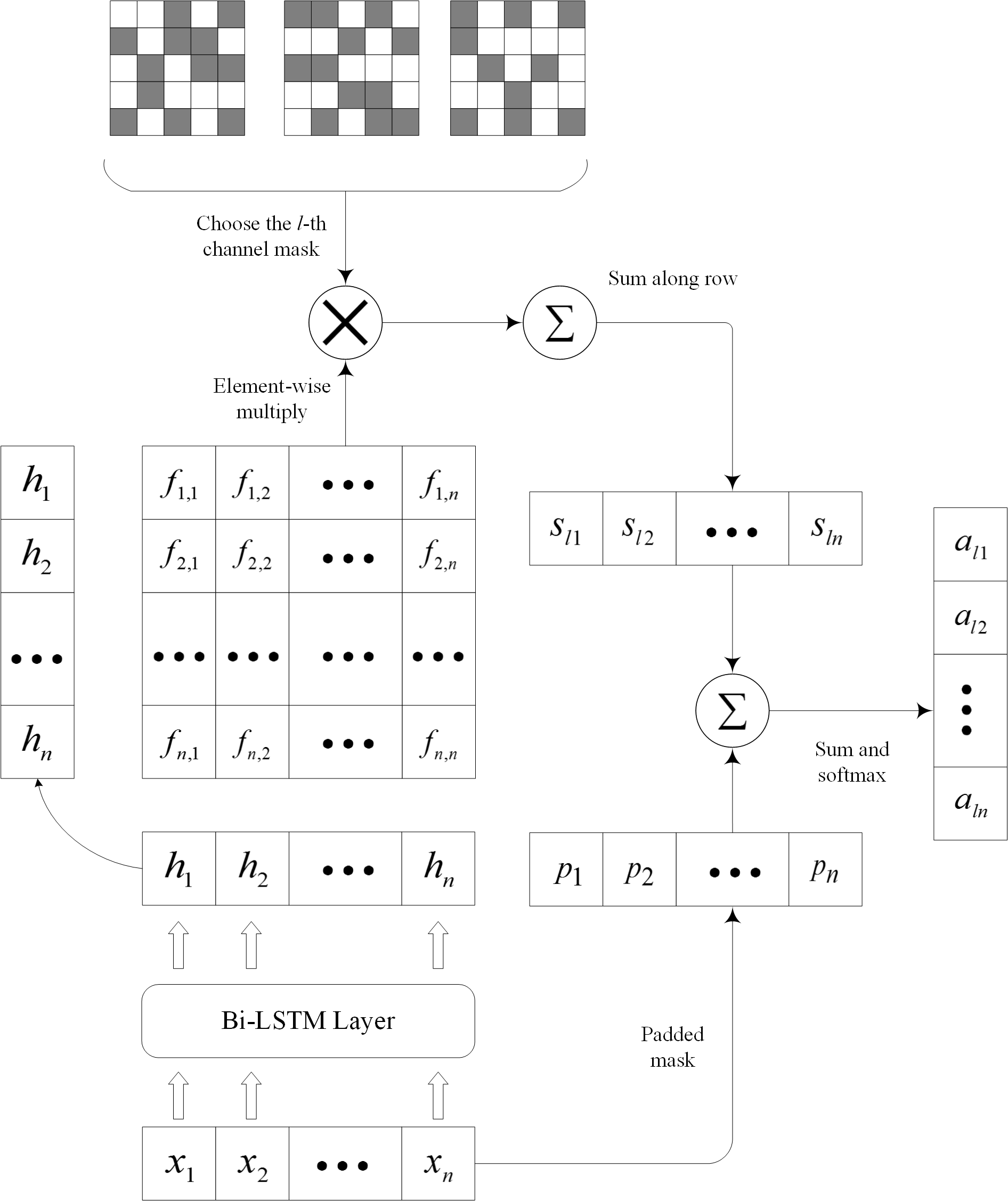}
\caption{syntactical attention mechanism}
\label{fig:syntactical}
\end{figure}

\begin{equation}
{c_{li}} = {a_{li}}{\cdot}{h_i},
\end{equation}

\begin{equation}
{C_l} = \left[ {{c_{l1}},{c_{l2}},{c_{l3}}, \cdot  \cdot  \cdot  \cdot  \cdot  \cdot ,{c_{ln}}} \right].
\end{equation}

Here, $ {c_{li}}$ denotes the new representation of $ h_i $ in the $l$-th channel and $ \otimes $ denotes element-wise product operation. The $ pad $ symbol still carries little information after
it is encoded by Bi-LSTM. So, if word $ x_k $ is a $ pad $ symbol, its syntactical attention $ s_{lk} $ will be subtracted from 99999 before softmax operation and so that $ a_{lk} $ will be close to 0 after
softmax. By concatenating all $C_{li}$, we obtain the $l$-th channel $ C_l $. The multichannel representations reflect the different contributions of different words to the semantics of a text, which is regarded as diversification of input information caused by data perturbation.

The whole process of the syntactical attention is shown in Fig.~\ref{fig:syntactical}.

\subsubsection{Semantical Attention Mechanism}

Given that a syntactical element (a word or a sentence) is encoded into an $ n $-dimensional vector $ {\left( {{v_1},{v_2},{v_3},......,{v_n}} \right)^T} $, each dimension in the embedding vector space corresponds to a specific latent semantic factor. Analyzing the different impacts of these semantic factors and selecting the informative ones can improve the performance of the downstream tasks.


Based on the above hypotheses, we propose the semantical attention mechanism to compute the semantical importance weight of each dimension in the input element:
\begin{equation}
\begin{split}
&\overrightarrow {scor{e_{li}}}  = {W_{l1}}^T\sigma \left( {{W_{l2}} \cdot {h_i} + {b_l}} \right), \\
&\mathop {{a_{li}}}\limits^ \to   = \frac{{\exp \left( {\overrightarrow {scor{e_{li}}} } \right)}}{{\sum\limits_i {\exp \left( {\overrightarrow {scor{e_{li}}} } \right)} }}, \\
&{c_{li}} = \mathop {{a_{li}}}\limits^ \to   \odot {h_{i}}, \\
&{C_l} = \left[ {{c_{l1}},{c_{l2}},{c_{l3}}, \cdot  \cdot  \cdot  \cdot  \cdot  \cdot ,{c_{ln}}} \right]. \\
\end{split}
\end{equation}
where $ c_{li} $ denotes the final representation of $ h_i $ in the $ l $-th channel. By concatenating all $c_{li}$ where $i \in \left[ {1,n} \right] $, we obtain the $ l $-th channel $ C_l $.

By combining the syntactical attention and the semantical attention,
multichannel is generated as follows:
\begin{equation}
\begin{split}
&{c_{li}} = {a_{li}}{\cdot}{(\mathop {{a_{li}}}\limits^ \to   \odot {h_{i}})}, \\
&{C_l} = \left[ {{c_{l1}},{c_{l2}},{c_{l3}}, \cdot  \cdot  \cdot  \cdot  \cdot  \cdot ,{c_{ln}}} \right]. \\
\end{split}
\end{equation}


\subsubsection{Convolutional Neural Network}

ConvNets utilize several sliding convolution filters to extract local features. Assume we have one
channel that is represented as
\begin{equation}
C = \left[ {{c_1},{c_2},{c_3},...,{c_n}} \right].
\end{equation}
Here, $ C \in {\mathbb{R}^{n \times k}} $, $ n $ is the length of
the input element, and $k$ is the embedded dimension of each input
element. In a convolution operation, a filter $ {\bf{m}} \in
\mathbb{R}^{lk} $ is involved in applying to consecutive $ l $ words
to generate a new feature:
\begin{equation}
{x_i} = f\left( {{\bf{m}} \cdot {{\bf{c}}_{i:i + l - 1}} + b} \right),
\end{equation}
where $ {\bf{c}}_{i:i + l - 1} $ is the concatenation of $ {c_i},...,{c_{i+l-1}} $. $ f $ is a non-liner activation function such as $ relu $ and $ b \in \mathbb{R} $ is a bias term. After the filter $ {\bf{m}} $ slide across $ \left\{ {{{\bf{c}}_{1:l}},{{\bf{c}}_{2:l + 1}},...,{{\bf{c}}_{n - l + 1:n}}} \right\} $, we obtain a feature map:

\begin{equation}\label{key}
{\bf{x}} = \left[ {{x_1},{x_2},...,{x_{n - l + 1}}} \right].
\end{equation}

We apply max-pooling operation over the feature map $ {\bf{x}} $ and
take the maximum value $ \hat x   = \max \{ {\bf{x}}\}  $ as the
final feature extracted by the filter $  {\bf{m}} $. This pooling
scheme is to capture the most dominating feature for each filter.
ConvNet obtains multiple features by utilizing multiple filters with
varied sizes. These features form a vector $ {\bf{r}} = \left[
{{x_1},{x_2},...,{x_s}} \right] $ ($s$ is the number of filters)
which will be passed to a fully connected softmax layer to output
the probability distribution over labels

\begin{equation}
y = softmax \left( {W \cdot {\bf{r}} + b} \right).
\end{equation}

Given a training sample ($ \textbf{x}^i  $, $ y^i $) where $ {y^i}
\in \left\{ {1,2, \cdots ,c} \right\} $ is the true label of $
\textbf{x}^i $ and the estimated probability of our model is $
\tilde y_j^i \in [0,1] $ for each label $ j \in \left\{ {1,2, \cdots
,c} \right\} $, and the error is defined as:

\begin{equation}
L({{\bf{x}}^i},{y^i}) = -\sum\limits_{j = 1}^c {if\{ {y^i} = j\} } \log (\tilde y_j^i).
\end{equation}
Here, $ c $ denotes the number of possible labels of $ \textbf{x}^i
$ and $ if\{ \dot \}  $ is an indicator function such that: $ if\{
{y^i} = j \}=1  $ if $ {y^i} = j $, $ if\{ {y^i} = j \}=0  $
otherwise. We employ stochastic gradient descent (SGD) to update the
model parameters and adopt Adam optimizer. Here, the ConvNet layer is intended to enhance the local N-gram coherence instead of merely averaging weighted sum, thus improving the discriminative ability to text classification.

\section{Experimental Study}
\label{sec:Experiment}

\subsection{Experiments Datasets}
We evaluate our model against other baseline models on a variety of datasets. Summary statistics of the
datasets are shown in Table \ref{tab:data}.
\begin{table}
    \centering
    \begin{tabular}{lcccccc}
        \toprule
        Data  & $ c $ & $ l $ & $ N $ & $ V $ & $ V_{word} $ & $ Test $  \\
        \noalign{\smallskip}\hline\noalign{\smallskip}
        MR & 2 & 20  & 10662 & 18765 & 16448 & CV \\
        Subj & 2 & 23  & 10000 & 21323 & 17913 & CV \\
        MPQA & 2 & 3  & 10606 & 6246 & 6083 & CV \\
        SST-1 & 5 & 18  & 11855 & 17836 & 16262 & 2210 \\
        SST-2 & 2 & 19  & 9613 & 16185 & 14838 & 1821 \\
        \bottomrule
    \end{tabular}
    \caption{Summary statistics of the datasets. $ c $: Number of classes. $ l $: Average length of sentences. $ N $: Size of datasets. $ V $: Vocabulary size. $ V_{word} $: Number of words present in the set of pre-trained word vectors, respectively. $ Test $: Size of test sets. $ CV $(cross validation): No standard train/test split and thus 10-fold CV was used. }
    \label{tab:data}
\end{table}

\begin{itemize}
    \item  \textbf{MR:} Short movie review dataset with one sentence per review. Each review was labeled with their overall sentiment polarity (positive or negative).

    \item \textbf{Subj:} Subjectivity dataset containing sentences labeled
    with respect to their subjectivity status (subjective or objective).

    \item \textbf{SST-1:} Stanford Sentiment Treebank—an
    extension of MR but with train/dev/test splits
    provided and fine-grained labels (very positive, positive, neutral, negative, very negative).

    \item \textbf{SST-2:} Same as SST-1 but with neutral reviews removed and binary labels

    \item \textbf{MPQA:} Opinion polarity detection subtask of the MPQA dataset.
\end{itemize}

\begin{table*}
    \centering
    \setlength{\tabcolsep}{7mm}{
        \begin{tabular}{lccccc}
            \toprule

            Model & MR & Subj & MPQA & SST-1 & SST-2 \\
            \noalign{\smallskip}\hline\noalign{\smallskip}
            Sent-Paser\cite{dong2015statistical} & 79.5 & -  & 86.3  & - & - \\

            NBSVM\cite{wang2012baselines} & 79.4 & 93.2  & 86.3 & - & - \\

            MNB\cite{wang2012baselines} & 79.0 & 93.6  & 86.3 & - & - \\

            F-Dropout\cite{wang2013fast} & 79.1 & 93.6  & 86.3 & - & - \\

            G-Dropout\cite{wang2013fast} & 79.0 & 93.4  & 86.1 & - & - \\

            \noalign{\smallskip}\hline\noalign{\smallskip}

            Paragraph-Vec\cite{le2014distributed} & - & -  & - & 48.7 & 87.8 \\

            RAE\cite{socher2011semi} & 77.7 & -  & - & 43.2 & 82.4  \\

            MV-RNN\cite{socher2012semantic} & 79.0 & -  & - & 44.4 & 82.9  \\

            RNTN\cite{socher2013recursive} & - & - & -& 45.7  & 85.4   \\

            DConvNet\cite{kalchbrenner2014convolutional} & - & - & -& \underline{48.5}  & 86.8   \\

            Fully Connected\cite{limsopatham2016modelling} & 81.59 & -  & - & - & -  \\

            ConvNet-non-static\cite{kim2014convolutional} & 81.5 & 93.4  & 89.5 & 48.0 & 87.2  \\

            ConvNet-multichannel \cite{kim2014convolutional} & 81.1 & 93.2 & 89.4 & 47.4 & \underline{88.1}  \\

            WkA+25\%fiexible\cite{lakshmana2016learning} & 80.02 & 92.68 & - & 46.11 & 84.29 \\

            Fully Connected \cite{limsopatham2016modelling} & 81.59 & - & - & - & - \\

            L-MConvNet \cite{guo2018integrated} & 82.4 & - & - & - & - \\

            Hclustering avg \cite{kim2018cnn} & 80.20 & - & - & - & - \\

            Kmeans centroid  \cite{kim2018cnn} & 80.21 & - & - & - & - \\

            \noalign{\smallskip}\hline\noalign{\smallskip}

            MahNN-1 & 82.17 & 92.96 & 89.61 & 47.02  & 86.43   \\
            MahNN-3 & \underline{82.57} & \underline{93.75} & \underline{89.75} & 47.58  & 86.85  \\
            MahNN-5 & 82.41 & 93.43 & 89.34 & 47.41  & 86.56   \\
            MahNN-7 & 82.23 & 93.36 & 89.46 & 47.16  & 86.29   \\
            MahNN-rv & 82.34 & 93.52 & 89.55 & 47.37  & 86.69   \\
            \bottomrule
    \end{tabular}}
    \caption{\small{Accuracies of MahNN against other models. We use underline to highlight wins. }}
    \label{tab:result}
\end{table*}

\subsection{Experiments Settings}

\begin{itemize}
    \item \textbf{Padding:} We first use $ len $ to denote the maximum length of the sentence in the training set. As the convolution layer requires input of fixed length, we pad each sentence that has a length less than $ len $ with $ UNK $ symbol which indicates the unknown word in front of the sentence. Sentences in the test dataset that are shorter than $ len $ are padded in the same way, but for sentences that have a length longer that $ len $, we just cut words at the end of these sentences to ensure all sentences have a length $ len $.

    \item \textbf{Initialization:} We use publicly available $word2vec$ vectors to initialize the words in the dataset. $ word2vec $ vectors are pre-trained on 100 billion words from Google News through an unsupervised neural language model.
    For words that are not present in the set of pre-trained words or rarely appear in data sets, we initialize each dimension from $ U\left[ { - 0.25,0.25} \right] $ to ensure all word vectors have the same variance. Word vectors are fine-tuning along with other parameters during the training process.

    \item \textbf{Hyper-parameters:} The feature representation of Bi-LSTM is controlled by the size of hidden states. We investigate our model with various hidden sizes and set the hidden size of unidirectional LSTM to be 100. We also investigate the impact of the size of the channels on our model. When the size of the channels is set to be 1, our model is a single channel network. When increasing the size of the channels, our model obtains a more semantic representation of the text. Convolutional filter decides the n-gram feature which directly influences the classification performance. We set the filter size based on different datasets and simply set the filter map to be 100. More details of hyper-parameters are shown on Table \ref{tab:set}.

\begin{table}
    \centering
    \begin{tabular}{lcc}
        \toprule
        Hyperparameter  &   Ranges  &  Adopt\\
        \noalign{\smallskip}\hline\noalign{\smallskip}
        $ Hidden\ size $ & \{16, 32, 50, 64, 100, 128, $ \cdots $\} & 100 \\
        $ L2 $ & \{0.0001, 0.0005, 0.001, 0.003, $ \cdots $\} & 0.0005\\
        $ Channel $ & \{1, 2, 3, 4, 5, 6, $ \cdots $\} &  3 \\
        $ Filter\ size $ & \{(2,3,4), (3,4,5), (4,5,6), $ \cdots $\} & -\\
        $ Filter\ map $ & \{10, 30, 50, 100, 150, $ \cdots $\} & 100\\
        \bottomrule
    \end{tabular}
    \caption{Hyper-parameters setting. $ Hidden\ size $: The dimension of unidirectional LSTM. $ L2 $: $ L2 $ regularization term. $ Channel $: The number of channels. $ Filter\ size $: The size of convolutional filters. $ Filter\ map $: The number of convolutional filter maps.}
    \label{tab:set}       
\end{table}

    \item  \textbf{Other settings:}  We only use one Bi-LSTM layer and one convolutional layer. Dropout is applied on the word embedding layer, the ConvNet input layer, and the penultimate layer. Weight vectors are constrained by $ L2 $ regularization  and the model is trained to minimize the cross-entropy loss of true labels and the predicted labels.

\end{itemize}

\begin{figure}[!t]
\centering
\includegraphics[width=\linewidth]{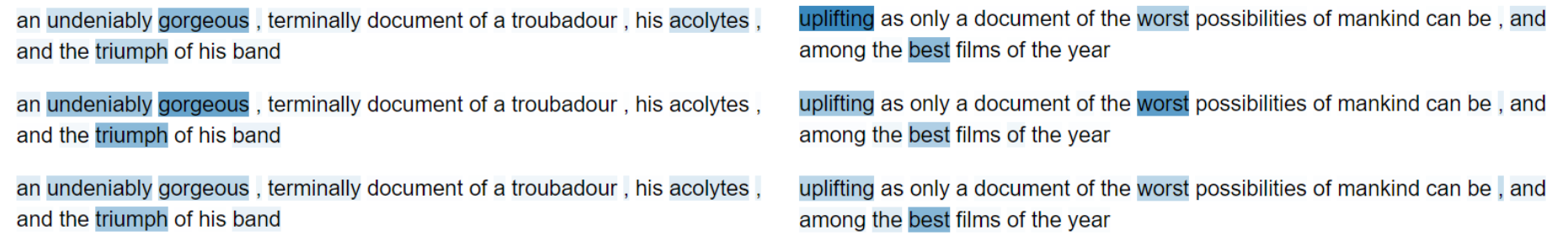}
\caption{Visualization of the syntactical attention weights learned by different channels}
\label{img}
\end{figure}

\subsection{Baselines}
We compare our model with several baseline methods which can be divided into the following categories:

\paragraph{$ \bf{Traditional\ Machine\ Learning} $} A statistical parsing framework was studied for sentence-level sentiment classification\cite{dong2015statistical}. Simple Naive Bayes (NB)
and Support Vector Machine (SVM) variants outperformed most
published results on sentiment analysis
datasets\cite{wang2012baselines}. It was shown in
\cite{wang2013fast} how to do fast dropout training by sampling from
or integrating a Gaussian approximation. These measures were
justified by the central limit theorem and empirical evidence, and
they resulted in an order of magnitude speedup and more stability.

\paragraph{$ \bf{Deep\ Learning} $} Word2vec \cite{le2014distributed} was extended with a new method called Paragraph-Vec, which is an unsupervised algorithm that learns fixed-length feature representations from variable-length pieces of texts, such as sentences, paragraphs, and documents.
Various recursive networks were extended
\cite{socher2012semantic,socher2011semi,socher2013recursive}.
Generic and target domain embeddings were incorporated to
ConvNet\cite{kalchbrenner2014convolutional}. A series of experiments
with ConvNets was trained on top of pre-trained word vectors for
sentence-level classification tasks \cite{kim2014convolutional}.
Desirable properties such as semantic coherence, attention mechanism
and kernel reusability in ConvNet were empirically studied for learning
sentence-level tasks \cite{lakshmana2016learning}. Both word
embeddings created from generic and target domain corpora were
utilized when it's difficult to find a domain corpus
\cite{limsopatham2016modelling}. A hybrid L-MConvNet model was proposed
to represent the semantics of sentences \cite{guo2018integrated}.
\begin{figure*}
    \centering
      \subfloat[]{\label{hidden}
       \includegraphics[width=0.45\linewidth]{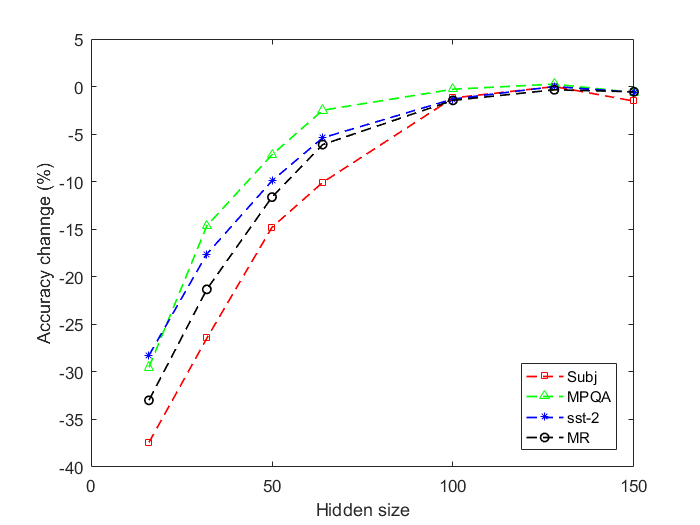}}
    \label{1a}\hfill
      \subfloat[]{\label{channel}
        \includegraphics[width=0.45\linewidth]{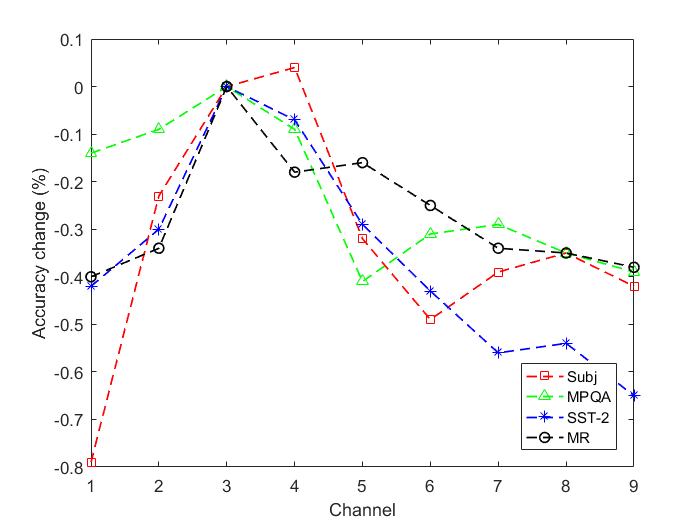}}
    \label{1b}\\
      \subfloat[]{\label{filter}
        \includegraphics[width=0.45\linewidth]{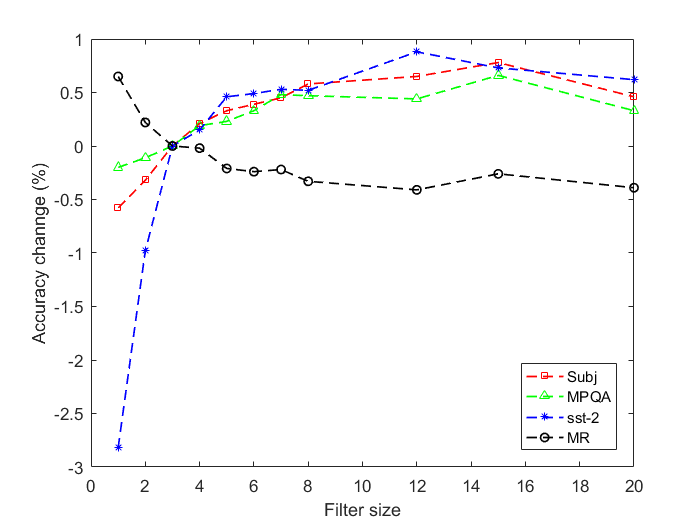}}
    \label{1c}\hfill
      \subfloat[]{\label{map}
        \includegraphics[width=0.45\linewidth]{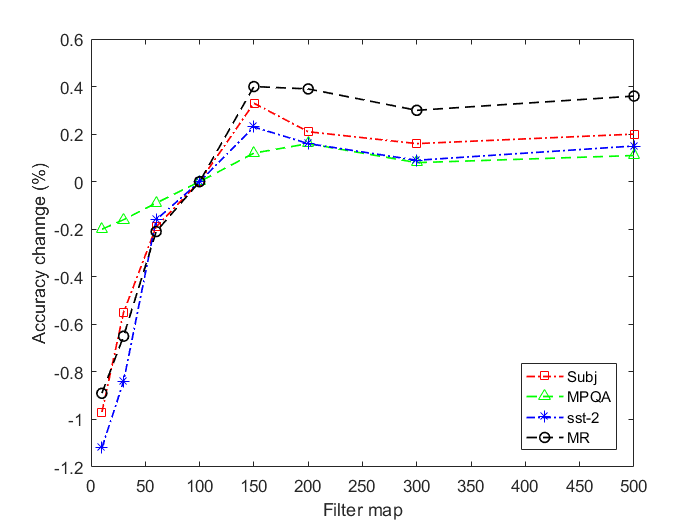}}
      \caption{Effect of hyper-parameters: (a) hidden size, (b) channel,  (c) filter size , and (d) the filter map.}
\end{figure*}

\subsection{Results and Analysis}

Table~\ref{tab:result} shows results of our model on five datasets against other methods. We refer to our model as MahNN-$ \{1,3,5,7, rv\} $, which stands for MahNN with different channel settings. As we can see from the Table~\ref{tab:result}, MahNNs exceed other models in 3 out of 5 tasks. For MR/Subj/MPQA, MahNN-$ 3 $ outperforms other baselines and we can get a rough observation that MahNN-3 performs better than MahNN-\{5,7\}, and they all perform better than MahNN-1, which is a single channel model. This phenomenon indicates that multichannel representation is effective, but continuing to increase the size of the channels does not improve our model all the time. We conjecture that it would be better to choose $ x $ according to the number of informative  words in the sentence. Take the following sentences for example:
\begin{enumerate}
    \item $ An $  $ undeniably $  $ gorgeous, $  $ terminally $  $ document $  $ of $  $ a $  $ troubadour, $  $ his $  $ acolytes, $  $ and $  $ the $  $ triumph $  $ of $  $ his $  $ band $.

    \item $ Uplifting $  $ as $ $  only $ $ a $ $ document $ $ of $ $ the $ $ worst $ $ possibilities $ $of$ $mankind$ $can$ $be$, $and$ $among$ $the$ $ best $ $films$ $of$ $the$ $year$.
\end{enumerate}

Fig.~\ref{img} shows the visualization of syntactical attention distribution of the above sentences.

The second sentence could not be labeled positive or negative without a  doubt if we focus on a single informative  word (``$uplifting$", ``$worst$" or ``$best$") alone. Only if these informative  words were all emphasized can this sentence be truly understood. ``$ Uplifting $" received more attention weight than other words in the first channel. ``$ worst $" received more attention weight in the second than the third channel and ``$ best $" received more attention weight in the third than the second channel. If the second channel is set to be an independent model, this sentence might be classified incorrectly. But MahNN-3 will still label this sentence as positive.
Multichannel essentially provides a way to represent a sentence from different views and provides diversification.

We also investigate the impact of the semantical attention on MahNN and find out that it considerably improves performance. MahNN-rv denotes MahNN-3 without applying the semantical attention mechanism. We owe the validity of the MahNN semantical attention mechanism to its selectivity of latent semantics that can better represent the texts in the specific given tasks. Actually, the semantical attention mechanism discriminates the perturbation of hidden states and makes the whole model more robust. Another advantage of the semantical attention mechanism is that it assigns different learning speeds to each dimension of the hidden state indirectly so that informative dimension could be tuned at a bigger pace than dimension of less information.

%

\subsection{Parameter Sensitivity}

We further evaluate how the parameters of MahNN impact its performance on the text classification task. In this experiment, we evaluate the effect of change of $ Hidden\ size $, $ Channel $, $ Filter\ size $, and $ Filter\ map $ on MahNN performance with other parameters remaining the same.

\begin{itemize}
    \item  \textbf{Impact of \textit{Hidden size}: }  Fig.\ref{hidden} shows the impact of \textit{Hidden size} on classification accuracy. It can be observed that the classification accuracy of the model increases with the increasing of hidden size. When the hidden size is set to be 128, the accuracy curve of the model tends to be flat or even begins to decline.
    So, the hidden size of Bi-LSTM affects the encoding of the document. If the \textit{Hidden size} is too small, it will lead to underfitting. If the \textit{Hidden size} is too large, it will lead to overfitting.

    \item  \textbf{Impact of \textit{Channel}: }   Fig.\ref{channel} shows the impact of \textit{Channel} on classification accuracy. We observe that the performance first rises and then tends to decline. When channel size is set to be 3, the model (MahNN-3) performs best on MPQA/SST-2/MR datasets. The model (MahNN-4) performs best on Subj dataset when channel size is set to be 4. This result shows that multichannel representations of texts help our model improve its performance. However, as the increasing number of the channels means the enlarged size of parameters, which might lead to overfitting.

    \item  \textbf{Impact of \textit{Filter size}: }  Fig.\ref{filter} shows the impact of \textit{Filter size} on classification accuracy. It can be observed that the optimal filter size settings of each dataset are different, and the accuracy curve of the MR dataset is opposite to the accuracy curve of other datasets. When \textit{Filter size} is between [10, 14], the model achieves high accuracy on MPQA/Subj/SST-2 datasets. But this performance improvement is not significant compared to the accuracy when \textit{Filter size} is 2.
    In order to reduce the size of the parameters, \textit{Filter size} of the model is set between [4, 8] in the experiment.

    \item  \textbf{Impact of \textit{Filter map}: }  Fig.\ref{map} shows the impact of \textit{Filter map} on classification accuracy. We can observe that the performance rises rapidly first and then tends to be flat.
    The number of \textit{Filter map} determines the number of feature maps generated after the convolution operation. Each feature map represents a certain feature of the text. The more the number of feature maps, the more features that the convolution operation can extract, and the accuracy of the model can be higher.
    But the number of features of the text is finite, and the increase in the number of \textit{Filter map} will also increase the size of trainable parameters,  which may lead to overfitting.
\end{itemize}

\section{Conclusion and Future Work}
\label{sec:Conclusion}

In this paper, we attempt to develop a hybrid architecture that can
extract different aspects of semantic information from the
linguistic data with diverse types of neural structures.
Intriguingly, we propose a novel hierarchical multi-granularity
attention mechanism, consisting of the syntactical attention at the
symbolic level and the semantical attention at the embedding level,
respectively. The experimental results show that the MahNN model
achieves impressive performances on a variety of benchmark datasets
for the text classification task. Moreover, visualization of
attention distribution illustrates that the hierarchical
multi-granularity attention mechanism is effective in capturing
informative semantics from different perspectives. We can draw the
following conclusions from our work:

\begin{enumerate}
    \item Hybrid neural architectures integrating a diversity of neural structures can improve the power of the representation learning from linguistic data. Richer semantic representations help to increase the capacity of deep understanding of texts and thus benefit to the downstream tasks in the NLP filed.

    \item Hierarchical multi-granularity attention mechanism plays a significant role in constructing the hybrid neural architecture. The fine-grained attention at the symbolic level can diversify the semantic representations of input texts and the coarser-grained attention at the latent semantical space enhance the local N-gram coherence for the following ConvNet layers, thus increasing the performance of the text classification tasks.
\end{enumerate}

There are several future directions to extend this work. First, we would investigate on applying a generative model to obtain multichannel representations of texts. Data augmentation in this way can represent much richer semantics. Second, ConvNets require the fixed-length inputs and perform some unnecessary convolution operations for NLP tasks. It is worthwhile to explore the novel ConvNet architecture processing with variable length. Moreover, we use simple calculating methods for the attention weights and this might not be able to demonstrate the full potential for the hierarchical multi-granularity attention mechanism. It would be intriguing to compute the attention weights with more advanced approaches such as transfer learning and reinforcement learning to further improve the performance.

\bibliographystyle{IEEEtran}
\bibliography{bibtex}

\end{document}